\documentclass{article}
\usepackage[nonatbib, final]{nips_2017}
\usepackage[utf8]{inputenc} 
\usepackage[T1]{fontenc}    
\usepackage{hyperref}       
\usepackage{url}            
\usepackage{booktabs}       
\usepackage{amsfonts}       
\usepackage{nicefrac}       
\usepackage{microtype}      
\usepackage{times}
\usepackage{epsfig}
\usepackage{graphicx}
\usepackage{amsmath}
\usepackage{amssymb}
\usepackage{multirow}
\usepackage{bbold}

\newcommand{\ba}{\begin{eqnarray}}
\newcommand{\ea}{\end{eqnarray}}

\newcommand{\bi}{\begin{itemize}}
\newcommand{\ei}{\end{itemize}}
\newcommand{\mycomment}[1]{}

\newcommand{\wdt}{.28}

\newcommand{\pos}{\mathbf{x}}
\newcommand{\posd}{\mathbf{x}'}
\newcommand{\posout}{\mathbf{x}_o}
\newcommand{\refeq}[1]{Eq.~\ref{#1}}

\newcommand{\dsp}{\mathbf{o}}
\newcommand{\xx}{\mathbf{x}}
\newcommand{\refsec}[1]{Sec.~\ref{#1}}
\newcommand{\todo}[1]{}
\newcommand{\reffig}[1]{Fig.~\ref{#1}}

\newcommand{\inp}{\pos}
\newcommand{\otp}{\pos_0}
\newcommand{\loss}{L}

\title{Mass Displacement Networks}
\author{
  Natalia Neverova \\
  Facebook AI Research\\
  Paris, France\\
  \texttt{nneverova@fb.com} \\
  \And
  Iasonas Kokkinos \\
  Facebook AI Research\\
  Paris, France\\
  \texttt{iasonask@fb.com}
}
\begin{document}
\maketitle
\begin{abstract}
Despite the large improvements in performance  attained by using deep learning in computer vision,  one can often further improve results with some additional post-processing that exploits the geometric nature of the underlying task. This commonly involves displacing the posterior distribution of 
a CNN in a way that makes it  more appropriate for the task at hand, e.g. better aligned with local image features, or more compact. 
In this work we integrate this geometric post-processing within a deep architecture,  introducing a differentiable and probabilistically sound counterpart to the common \textit{geometric voting} technique  used for evidence accumulation in vision.  We refer to the resulting neural models as  Mass Displacement Networks (MDNs), and apply them to human pose estimation in two distinct setups: (a) landmark localization, where we collapse a distribution to a point, allowing for precise localization  of body keypoints and (b) communication across body parts, where we transfer evidence from one part to the other,  allowing for a globally consistent pose estimate. We evaluate on large-scale pose estimation benchmarks, such as MPII Human Pose and COCO datasets,
and report systematic improvements when compared to strong baselines.

\end{abstract}
\section{Introduction}
\label{introduction}
The advent of deep learning has reduced the amount of hand-engineered processing required for computer vision by integrating many  operations such as pooling, normalization, and  resampling within Convolutional Neural Networks (CNN). The succession of such operations gradually discards  the effects of irrelevant signal transformations, allowing  
the higher layers of CNNs to exhibit increased robustness to small input perturbations.  While this invariance is  desirable for high-level vision tasks, it can harm tasks such as pose estimation where one aims at precise spatial localization, rather than abstraction.  

It is therefore  common to apply some form of computer vision-based post-processing on top of CNN-based scores to obtain sharp, localized geometric features. 
One of the first steps in this direction has been the use of structured prediction on top of semantic segmentation,  e.g. by combining image-based DenseCRF~\cite{Koltun13} inference with CNNs for semantic segmentation~\cite{ChenPKMY14}, training both systems jointly~\cite{crfrnn}, or more recently learning  CNN-based pairwise terms in structured prediction modules~\cite{Chandra16,Hengel16}. All of these works involve coupling decisions so as to reach some consistency in the labeling of global structures, typically coming in the form of smoothness constraints. While this is meaningful for tasks where information is spread out, such as semantic segmentation, we are interested in more general transformations, some of which are illustrated in \reffig{fig:teaser}. For instance, we consider tasks that require outputs in the forms of 1-D or 0-D outputs (boundary and keypoint detection, respectively), effectively collapsing the spatially extended  output of a CNN into lower-dimensional structures. Even though in principle this could be cast in structured prediction terms, the resulting optimization problem amounts to maximizing a submodular function \cite{Blaschko11} and can only be approximately optimized. We therefore turn  to  geometry-based, rather than optimization-based methods, and pursue their  incorporation in the context of deep  learning. 

Our starting point is the understanding that requiring  high spatial accuracy from a purely CNN-based deep architecture is misusing the network's abilities: by design, the CNN feature maps get increasingly smooth as we go deeper. We can instead combine these smooth CNN-based classification results  with an equally smooth \textit{displacement field} obtained from another CNN branch, indicating to every pixel where its mass should be displaced. This is achieved by separately predicting values of x- and y- components of the displacement vectors (or \textit{offsets}) of all pixels.
Even though the displacement field may be smooth, if its value is accurate, then result can become sharp -- in \reffig{fig:teaser} we are displaying some indicative examples of a smooth response being manipulated by smooth displacement fields that turn it into quite different shapes, that could be appropriate for a variety of visual  tasks. 

\renewcommand{\wdt}{.14}
\newcommand{\wda}{.07\linewidth}
\newcommand{\wdi}{.131\linewidth}
\begin{figure}
\begin{center}
\includegraphics[width=\linewidth]{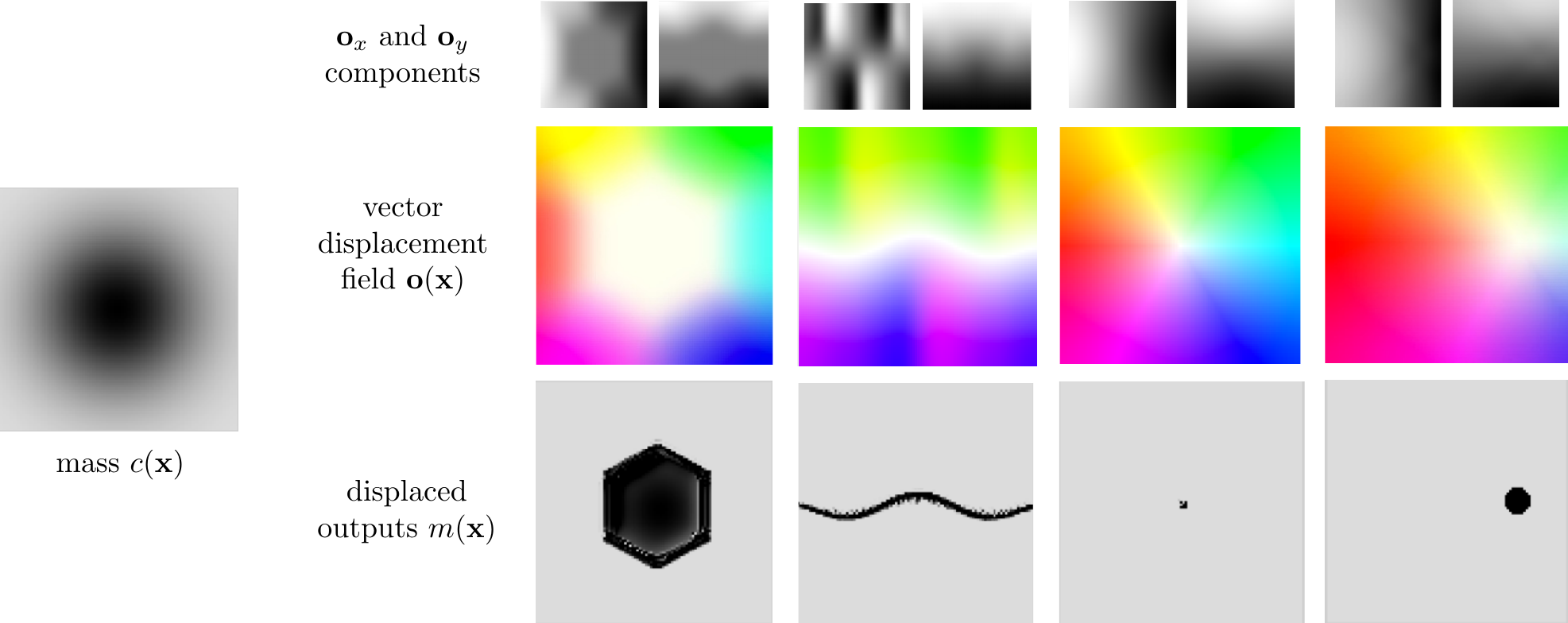}\hfill
\end{center}
\hspace{56mm} (a)\hspace{\wdi} (b)\hspace{\wdi} (c)\hspace{\wdi} (d)
\caption{
{\small
The low spatial resolution of CNNs results in overly smooth per-pixel confidence scores (\textit{mass}), as shown in the image on the left. Rather than stretch the CNN's capabilities in order to obtain spatially sharp responses, we propose instead to append a \textit{dispacement field} as another CNN output that rearranges the classification scores, lending more evidence to the ground truth positions. The ${\mathbf{o}}_x$ and ${\mathbf{o}}_y$-components of different displacement fields $\mathbf{o}(\mathbf{x})$ are shown in the top row on the right (the middle row shows the same components presented as a vector field and displayed in color, for illustrative purposes).
These are combined by a Mass Displacement module into a sharp decision, shown in the bottom row. This can amount to making the classification obtain a particular shape, e.g. through alignment with image boundaries (a), a 1D structure such as a line, or a curve (b), a point (c), or displacing to another position (d). While both the raw network outputs (mass) and the displacement fields are smooth, the final results are sharp. }}
\label{fig:teaser}
\end{figure}

What we are proposing can be understood as \textit{reinventing} geometric voting in the context of  deep learning: in a host of  computer vision tasks \cite{Ballard81,LeibeLS08,MajiM09,GallYRGL11,RazaviGKG12,BarinovaLK12} voting can be used to first associate an observation with positions that it supports and then shortlist structures that are supported by multiple observations, 
 e.g. many points voting for a line or a cycle \cite{Ballard81}, object parts voting for an object's 2D  \cite{LeibeLS08} or 3D pose \cite{ism3d}, or many object hypotheses voting for a single object bounding box \cite{GidarisK15}. Our work was actually motivated by the recent success of such  schemes for landmark localization in \cite{gpapan},  instance segmentation in \cite{WuSH16b}, and bounding box post-processing in \cite{GidarisK15}. 
 
All of these approaches however are plagued by the heuristic nature of geometric voting, that makes them only applicable as post-processing steps. 
For example in \cite{gpapan} posterior probabilities are being displaced and then accumulated  which  results in score maps that can be larger than one -- disqualifying them from training with losses appropriate for classification.  The authors end up using the cross-entropy loss for the original CNN and the L2 loss for the second stage, while also not training the displacement fields end-to-end -- as such it is unclear if the displacement fields are really pointing to the positions that they should. Instead, in this work we develop this somehow ad-hoc  post-processing  into a module that can easily be combined with existing architectures and  trained end-to-end. 
  
In particular, we treat geometric voting  as a \textit{differentiable} operation,  allowing us to train the CNN-based score maps and displacement fields in an end-to-end manner, ensuring that both arguments to the voting function are optimizing the final system's performance. Each displaced point is dilated by a kernel to support a region around its novel position, and
in the output space every position accumulates evidence from  input points that can support  it.  The currently common approach of adding the posterior probabilities is also not  justified probabilistically \cite{Williams}. Instead we consider a probabilistically sound method of accumulating evidence that forces the final outcomes to stay smaller than one -- as such the output of our operation lends itself to training with probabilistic criteria, such as the cross-entropy loss.  

Since our approach combines spatial transformation with the geometric manipulation of a probability mass function, we call a network incorporating our method a Mass Displacement Network (MDN).
We explore two tasks: (i) human body landmark localization through \textit{within-part} voting, where the coarse score map of a part is sharpened by a voting process (ii) human pose estimation through \textit{across-part}  voting, where every body part score map votes for the presence of other parts. 
We provide systematic demonstrations of improvements achieved by MDNs over strong baselines on large-scale benchmarks in human pose estimation both in single person (MPII Human Pose dataset) and multi-person (COCO dataset) setups. \medskip

  
 \textbf{Connections to other works:}
 Apart from the works mentioned already, our approach  has connections to Spatial Transformer Networks (STNs), introduced in  \cite{JaderbergSZK15} to bring raw images into correspondence and remove intra-class variation that can be modelled in terms of image deformations. 
The tacit assumption underlying STNs is that the input and output fields are related by a diffeomorphic transformation, such as a similarity transformation or an affine map, meaning that the dimensionality of structures is preserved. Instead, here we consider transformations that allow us to collapse 2D structures into lower-dimensional structures, such as points, or lines. 
Furthermore, STNs typically consider a single global parametric transformation, while we have a non-parametric transformation determined by a fully convolutional layer. Finally, as we explain in \refsec{voting}, STNs are designed  like image interpolation operations, and are typically used at the input of a network, while we cater for evidence accumulation, and our module is intended to be appended at the end of a network, or generally after some decisions have been produced by a CNN.

In a work parallel to our own \cite{daiactive}, the authors have introduced active CNNs,  which allow a neuron to pick incoming neurons from  input positions determined dynamically through a CNN-based deformation. While this work shares with us the idea of using a convolutional, CNN-based deformation field,  in our case we have input neurons deciding where they move to, rather than output neurons deciding from where to pool their information. As such our approach seems to be better suited for collapsing densities and accumulating evidence to certain positions, while the work of \cite{daiactive} seems better suited for the task of discarding the effect of deformations.

\begin{figure}
\begin{center}
\includegraphics[width=\linewidth]{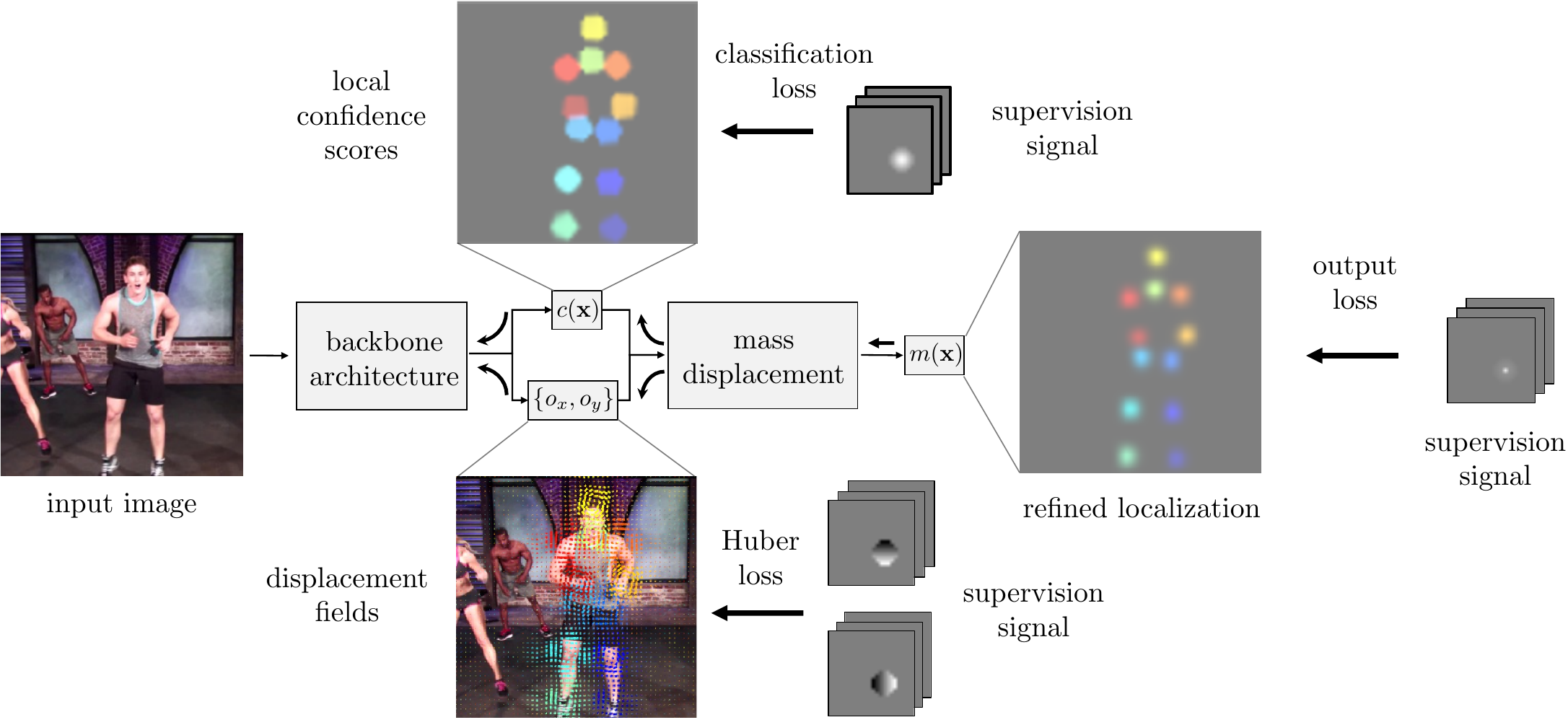}
\caption{%
{\small Architecture of a Mass Displacement Network (MDN): the convolutional layers of a CNN are trained with loss functions that allow for some uncertainty in the  localization of landmarks, accommodating  their inherently smooth responses.    A voting operation  combines these and collapses the smooth CNN predictions into 
sharp landmarks. We  treat the voting mechanism as a differentiable module and use it for end-to-end training.}}
\label{fig:mtn}
\end{center}
\end{figure}

\section{Mass Displacement Networks}
\label{mtn}
We start by describing in \refsec{sec:vot} the non-probabilistic, geometric voting process currently employed in recent works
\cite{gpapan,Hengel16} and then propose a principled variant that relies on the noisy-or rule \cite{Pear88} allowing us to use the cross-entropy loss during training. We then turn in \refsec{sec:gradient} to the equations used for end-to-end training of the resulting Mass Displacement Network. 

\label{voting}
\subsection{Additive and Noisy-OR Voting}
\label{sec:vot}
We consider that both our local evidence functions and the output structures reside in a two-dimensional space.
In particular we consider that for any position $\pos{=}(x,y)$ a convolutional network provides us with two outputs: firstly 
an estimate of the local confidence for the presence of a feature $c(\pos)$, and  secondly  an estimate of the predicted structure's position.
The latter is expressed as an horizontal/vertical displacement $\mathbf{o}(\pos)$ (or, \textit{offset}) that should be applied to the current position to obtain the refined estimate $\posd$:
\ba 
\posd(\pos) = \pos + \dsp(\pos)  = (x,y) + (o_x(x,y),o_y(x,y)). \nonumber
\ea
For landmark localization in within-part voting the displacement field can act like 
a residual correction signal, while for across-part voting it reflects relative part locations. 
 We can accommodate  spatial uncertainty in the predicted position by supporting  a structure not only at $\posd$, but also in the vicinity of the same point. This can be accomplished by dilating the local confidence $c(\pos)$ with a kernel, e.g. $K(\posout-\posd) = \exp\left(-\frac{\|\posout-\posd\|^2}{2\sigma^2}\right)$,
that allows us to smoothly decrease our support as we move further away from $\posd$. 
Combining evidence from multiple points is typically done through summation:
\ba
m(\posout) = \sum_{\pos} K(\posout  - \left[\pos + \dsp(\pos)\right]) c(\pos), \label{eq:voting}
\ea
where for every output position $\posout$ we sum the support delivered by  all input positions $\pos$.
This has been the setting used for instance in \cite{gpapan} and \cite{Hengel16} 
for landmark localization and instance segmentation, respectively.
In these works a CNN is trained with a cross-entropy loss for $c(\pos)$ and a regression loss for $\dsp(\pos)$, while \refeq{eq:voting} is used at test time to deliver more accurate estimates of the desired structures.

The  operation in \refeq{eq:voting} can be justified in the context of image interpolation, as in the case of Spatial Transformer Networks \cite{JaderbergSZK15}, or in standard Kernel Density Estimation (KDE), but not  as a method of accumulating evidence \cite{Williams}. 
The main problem, detailed in Appendix A, is that we cannot simultaneously guarantee that the input and output fields both lie in $[0,1]$, so that they can be trained with the cross-entropy loss, and that a confident posterior at $\pos$ will confidently support its displaced replica at $\pos + \dsp(\pos)$, i.e. $K(\mathbf{0}){=}1$. 

We can  guarantee both requirements by replacing summation with maximization (i.e. perform a \lq\lq Transformed-Max-Pooling operation\rq\rq). Our experiments with this approach were underwhelming, understandably because we do not really accumulate evidence from many points, but rather rely on the single most confident one. Instead we propose to use differentiable approximations of the maximum operation \cite{Pear88,ViolaPZ05,babenko} that allow us to softly combine multiple pieces of evidence while ensuring that the outputs are probabilistically valid.

In particular we use the  noisy-or combination rule \cite{Pear88} which provides a probabilistic counterpart to a logical OR-ing operation. We consider that we have $J$ pieces of evidence about the presence of a feature, each being true with a probability of $p_j, j{=}1{\ldots}J$. The  noisy-or  operation expresses the probability $p$ of the presence of the feature as follows:
$1{-}p{=}\prod_{j=1}^J (1{-}p_j)$,
namely the feature is absent if all supporting pieces of evidence are simultaneously absent -- as such, any additional piece of evidence can only increase the estimated value of $p$. 
If now we replace $p_j$ in the above formula with $K(\posout  - \left[\pos + \dsp(\pos)\right]) c(\pos)$ we obtain the following rule 
for combining evidence in the MDN:
\ba
m(\posout) = 1- \prod_{\pos} \left[1- K(\posout  - \left[\pos + \dsp(\pos)\right]) c(\pos)\right]. \label{eq:noisy-or}
\ea
We note that we can use a first-order approximation to obtain \refeq{eq:voting} from \refeq{eq:noisy-or} if all of the individual terms $K(\posout{-}\left[\pos{+}\dsp(\pos)\right]) c(\pos)$ are very small, which in hindsight gives some explanation for the practical success of \refeq{eq:voting}. However, in \refeq{eq:noisy-or} we have $m(\posout){\in}[0,1]$  which  allows us to use the cross-entropy loss throughout training, by virtue of being probabilistically meaningful. Our experiments show that this yields equally good results as the currently broadly used  heuristic of regressing to Gaussian functions \cite{TompsonGJLB15,NewellYD16,BulatT16}, while being simpler and cleaner.

\subsection{Back-propagation through an MDN module}
\label{sec:gradient} 

The input-output mapping defined by \refeq{eq:voting} is differentiable with respect to both input functions,~$\dsp(\pos),c(\pos)$, and as such lends itself to end-to-end training with back-propagation. 
Given a gradient signal $\delta(\cdot){=}\frac{\partial \loss}{\partial m(\cdot)}$ that  dictates how the output layer activations should change to decrease the network loss $\loss$, we obtain the update equations for $c(\cdot)$ and $\dsp(\cdot){=} (o_x(\cdot),o_y(\cdot))$ through the following chain rule:
\ba  
\frac{\partial \loss }{\partial c(\inp)}
= \sum_{\otp} \delta_{\otp}
\frac{\partial m(\otp)}{\partial c(\inp)}, \quad 
\frac{\partial \loss }{\partial \{o_x/o_y\}(\inp)} = \sum_{\otp} \delta_{\otp}
\frac{\partial m(\otp)}{\partial \{o_x/o_y\}(\inp)},
\label{eq:bprop}
\ea
where 
the summation runs over the  top-layer neurons $\otp$ that send gradients back to neuron $\inp$.
Turning to the computation of the partial derivatives in
\refeq{eq:bprop}, the use of displacement fields means that we no longer have a standard convolutional layer; 
an input position $\inp$ can potentially influence 
any other output position $\otp$, as dictated by 
\refeq{eq:noisy-or}. 
For convenience we rewrite \refeq{eq:noisy-or} as follows:
\ba 
m(\otp) = 1 - \prod_{\inp} \big(1 - w(\inp,\otp) c(\inp)\big), \quad \mathrm{where}~~w(\inp,\otp) = K\big(\otp - [\inp + \dsp(\inp)]\big),
\ea
indicates the amount by which $\inp$ influences $\otp$.
Using the same steps as in \cite{ViolaPZ05}, in case of a Gaussian kernel $K$ we have:
\ba
\frac{\partial m(\otp) }{\partial c(\inp)} =
  w(\inp,\otp) \frac{1 \!-\! m({\otp})}{ 1 \!-\! w({\inp}) c({\inp})},
\quad \frac{\partial m({\otp}) }{\partial o_x(\inp)} \!=\! 
\frac{\partial m({\otp}) }{\partial w({\inp,\otp})}
K'\big(\otp\!-\![\inp\! + \!\dsp(\inp)]\big)\big[x_0\! - [x\! + o_x(\inp)]\big]
\nonumber
\ea 
where $x_0,x,o_x(\inp)$ are the horizontal components of $\otp,\inp,\dsp[\inp]$ respectively. 

\begin{figure*}
\begin{center}
\includegraphics[width=0.95\linewidth]{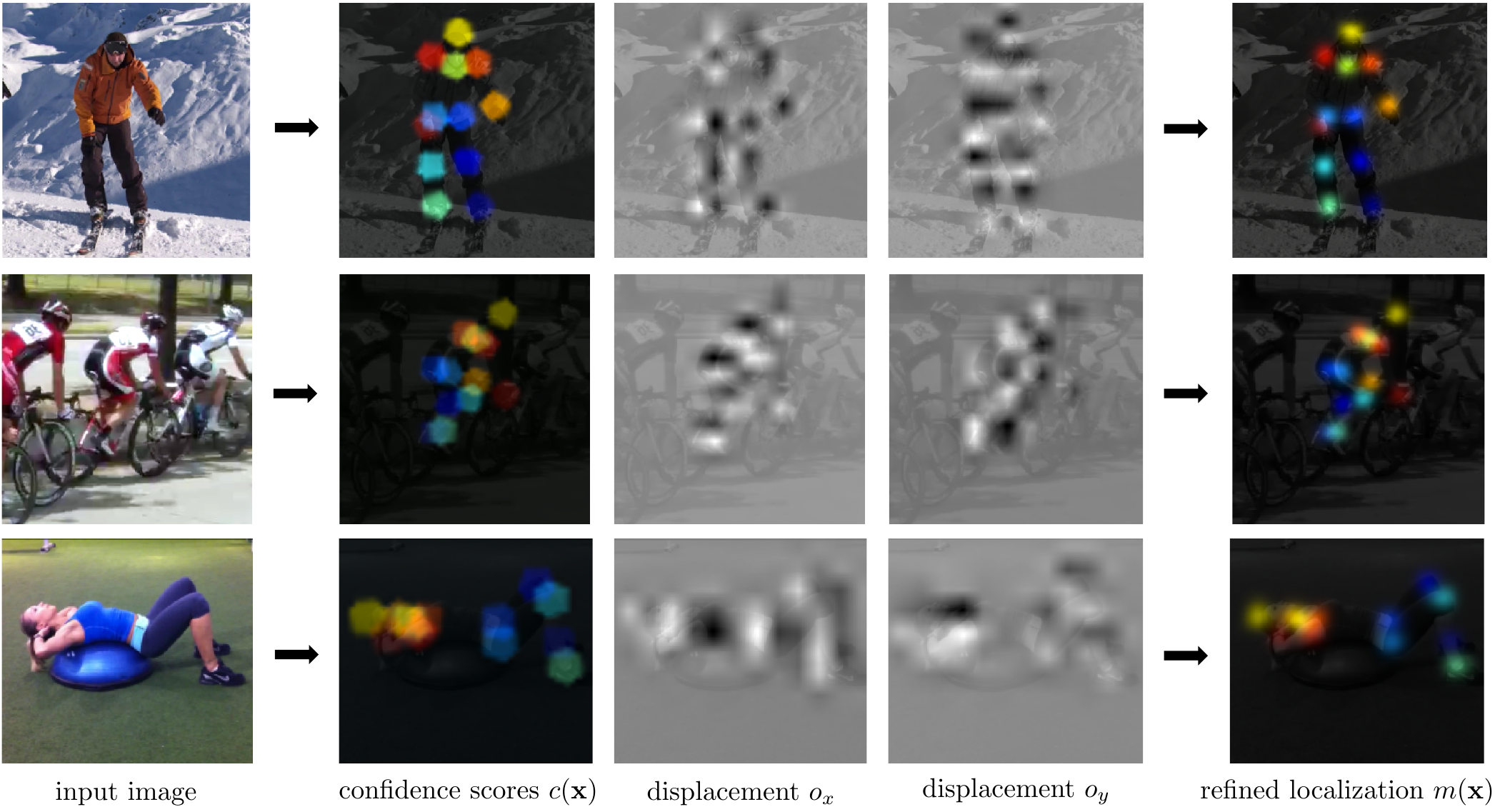}
\caption{\small MDN computation in practice: when presented with an image, the three convolutional branches of our network deliver the smooth posterior probabilities $c(\pos)$ and horizontal and vertical offsets $o_x$ and $o_y$ shown in the middle row (for simplicity, for every kind of output here we display a sum over all planes corresponding to different keypoints). The MDN combines these into the sharper joint estimates $m(\pos)$ shown on the right.}
\label{fig:localresults}
\end{center}
\end{figure*}

\section{Experimental Evaluation}
We present experiments in two setups: firstly, \emph{single-person} pose estimation on the MPII Human Pose dataset \cite{AndrilukaPGB14}, where the position and scale of a human is considered known in advance. This disentangles the performance of the pose estimation and object detection systems. Secondly, we consider \emph{human pose estimation \lq\lq in-the-wild\rq\rq} on the COCO dataset \cite{mscoco}, where one needs to jointly tackle detection and pose estimation. We use different baselines for both setups, since there is no common strong baseline for both. In both cases  MDNs  systematically improve  strong baselines. 

\begin{table*}[!t]
  \begin{center}
  \caption{\small Relative performance of the MDN applied to isolated landmarks and trained with different combination rules. All models are based on ResNet-152 and are tested on the validation set of MPII Single person. The third baseline is obtained by applying a max operation instead of a sum (additive MD) or a product (noisyOR). $k_f$ denotes the kernel size.}\vspace*{3mm}\small
  \label{tbl:local}
    \begin{tabular}{l|c|c|ccccccc}
      \toprule
     \multirow{2}{*}{\hspace*{9mm} Model } 
&\multirow{2}{*}{\!\!\!\!\begin{tabular}{l} \hspace*{5pt}No\\voting \end{tabular}\!\!\!\!}
&\multirow{2}{*}{\!\!\!\!\begin{tabular}{l} Bilinear\\\hspace*{5pt}kernel \end{tabular}\!\!\!\!}
& \multicolumn{6}{c}{Gaussian kernel}\\ \cmidrule{4-9}
     &&& $k_f{=}3$\! & $k_f{=}5$\! & 
     $k_f{=}7$\! & $k_f{=}9$\! &
     \!$k_f{=}11$\! & \!$k_f{=}13$\!\\ \midrule
Baseline, additive & \multirow{4}{*}{84.31} & 87.54 & 87.70\! & 88.01\! & 88.11\! & 88.15\! & 88.19 & 88.19 \\
Baseline, noisyOR && 87.49 & 87.63\! & 87.84\! & 87.98\! & 88.08\! & 88.19 & 88.11\\
Baseline, max && 86.69 & 86.38 & 86.22 & 86.03 & $85.96$ & 85.34& 85.12\\
Spatial Transformer  \cite{JaderbergSZK15} & & 88.28 &  &  &  &  &  & \\
\midrule
MDN-additive &  & \textbf{88.60} & $\times$ & \textbf{88.63}\! & $\times$ & $\times$ & \textbf{88.61}\!& $\times$  \\
MDN-noisyOR &  & \textbf{88.61} & $\times$ & \textbf{88.58}\! & $\times$ & $\times$ & \textbf{88.32} & $\times$\\
\bottomrule
  \end{tabular}
  \end{center}
\end{table*}

\begin{table*}[!t]
  \begin{center}
  \caption{\small Ablation of interplay between MDN and network architecture choices (PCKh on MPII-val). }\vspace*{3mm}\small
  \label{tbl:resnets}
    \begin{tabular}{l|cccc}
      \toprule
     \hfill Model \hfill\, & Resnet-50 & Resnet-101 & Resnet-152 & Hourglass-8\\
\midrule
Baseline, no voting & 83.29 & 84.28 & 84.31 & 89.24\\
Baseline, additive, bilinear kernel & 86.50 & 87.50 & 87.54 & 89.43\\
Baseline, additive, Gaussian $k_f{=}11$ & 87.17 & 88.10 & 88.19 & 89.70 \\
Baseline, noisyOR, bilinear kernel & 86.42 & 87.46 & 87.49 & 89.49 \\
Baseline, noisyOR, Gaussian $k_f{=}11$ & 87.12 & 88.08 & 88.11 & 89.67  \\
\midrule
MDN-additive, bilinear kernel  & 87.23 & 88.42 & {88.60} & \textbf{89.72}\\
MDN-noisyOR, bilinear kernel  & \textbf{87.25} & \textbf{88.52} & \textbf{88.61} & {89.64}\\
\bottomrule
  \end{tabular}
  \end{center}
\end{table*}

\subsection{Single person pose estimation}

\subsubsection{Experimental setup}
\textbf{Dataset \& Evaluation:} We evaluate several variants of MDNs on the MPII Human Pose dataset \cite{AndrilukaPGB14} which consists of 25K images containing over 40K people with annotated body joints. We follow the \textit{single person} evaluation protocol, i.e. use a subset of the data with isolated people assuming their positions and corresponding scales to be known at test time.
We follow the standard evaluation procedure of \cite{AndrilukaPGB14} and report  performance with the common Percentage Correct Keypoints-w.r.t.-head (PCKh) metric \cite{YangR13}. As in \cite{BulatT16,NewellYD16}, we refine the  test joint positions by averaging network predictions obtained with the original and horizontally flipped images.

\textbf{Implementation:} We conduct the first exhaustive set of experiments by fine-tuning  ImageNet-pretrained ResNet architectures  \cite{HeZRS15}. 
We  substitute the output linear layer and the average pooling that precedes it  with a \textit{bottleneck} convolution layer of spatial support $1{\times}1$ that projects its 2048-dimensional input down to  512 dimensions. This acts like a \textit{buffer layer} between the pretrained network and the  pose-specific output layers. As in \cite{gpapan}, we reduce the amount of spatial downsampling in such networks by reducing the stride of the first residual module in conv5 block from 2 or 1, and  employ atrous convolutions afterwards \cite{ChenPKMY14}. 
As a result, the network takes as an input a cropped image of size $256{\times}256$, produces a set of feature planes with spatial resolution of $16{\times}16$ (rather than $8{\times}8$). These are then bilinearly upsampled to produce the outputs of size $64{\times}64$. \\
On top of this common network trunk operate three convolutional branches that deliver the three inputs of the MDN, namely confidence $\c(\pos)$ and displacement fields $o_x,\,o_y$. Each such branch is a single convolutional layer of spatial support $1{\times}1$ which maps the $512$ feature planes to $N{=}16$ dimensions, where $N$ is the number of landmarks to be localized. The outputs of these branches are passed to the MD layer, which in turn outputs the final refined localizations at the same resolution. 

We also present preliminary  experimental results with hourglass networks \cite{NewellYD16}, that have even higher performance on MPII -- we apply a similar re-purposing as the one outlined above by introducing additional convolutional heads for predicting displacement fields after each stack of the network (where the final estimates for the offsets are obtained by taking a sum over predictions at each step).


\textbf{Training:} 
In these experiments, we test the perfomance of both additive and noisyOR MDNs. We train the network with three kinds of supervision signals applied to the following outputs:\\
(a) the confidence maps $c(\mathbf{x})$ trained with pixelwise \textit{binary cross entropy loss}. The supervision signal at each point $(x_i^{(j)},y_i^{(j)})$ from output plane $j$ is formulated in the form of binary disks centered at each keypoint location: $\hat{c} = \mathbb{1}[|x^{(j)}_{i}{-}x_0^{(j)}|\leqslant\varepsilon_c]$, where $(x_0^{(j)}\!,y_0^{(j)})$ is the ground truth position of joint $j$, $\varepsilon_c{=}4$.\\
(b) two offset planes $\{o_x,\,o_y\}$ learned with \textit{robust Huber loss} applied solely in the $\varepsilon_c$-vicinity of the ground truth position of every keypoint. The ground truth value for each point $(x_i^{(j)},y_i^{(j)})$ in the $\varepsilon_c$-vicinity of joint $j$ voting for joint $k$ is defined as follows:
\begin{equation}
\label{eq:offsets}
\hat{o}_{x,i}^{(j,k)}{=} \frac{(x^{(j)}_{i}{-}x_0^{(k)})}{d}[|x^{(j)}_{i}{-}x_0^{(j)}|\leqslant\varepsilon_c],
\end{equation}
 where, as before, $(x_0^{(m)}\!,y_0^{(m)})$ is the ground truth position of joint $m$ and $d$ is a normalization factor (defined below). The vertical component $\hat{o}_y$ is defined analogously. \\
(c) the final refined localizations $m(\mathbf{x})$. In this case, depending on the aggregation rule, we apply either \textit{MSE regression loss} (for additive mass displacement) or \textit{binary cross entropy loss} (in case of noisyOR aggregation). The final supervision signal is formulated in the form of a Gaussian (additive MDN) or a binary disk (noisyOR MDN) in the same way as in (a) but with a smaller value of $\varepsilon_m{=}1$.\\
We would like to note here that supervising the network with a single loss (c) is possible and produces similar final results but at cost of significantly slower convergence.

All networks are trained using the training set of MPII Single person dataset with artificial data augmentation in the form of flipping, scaling and rotation, as described in \cite{NewellYD16}. We employ the RMSProp update rule, initial learning rate 0.0025, learning rate decay 0.99, and as in  \cite{NewellYD16} use a validation set of 3000 heldout images for our ablation study. 

We perform evaluation on two separate tasks of within-part and cross-part voting:\\
(a) \textbf{local mass displacement(within-part voting):}
in this setting, the offset branches receive their supervision signal in the form of local distributions of horizontal and vertical offsets defined as in \ref{eq:offsets}, where $j{=}k$ and $d{=}\varepsilon_c{-}1$;\\
(b) \textbf{global mass displacement (cross-part voting):} the implementation of the cross voting mechanism is similar to the previous case, but $j{\neq}k$ and $d{=}X$, where $X{\times}X$ is the output resolution.
In this case, we found it more effective to restrict connectivity between joints and perform cross-joint voting along the kinematic tree starting from the center of the body.


\begin{figure*}[!t]
\begin{center}
\includegraphics[width=\linewidth]{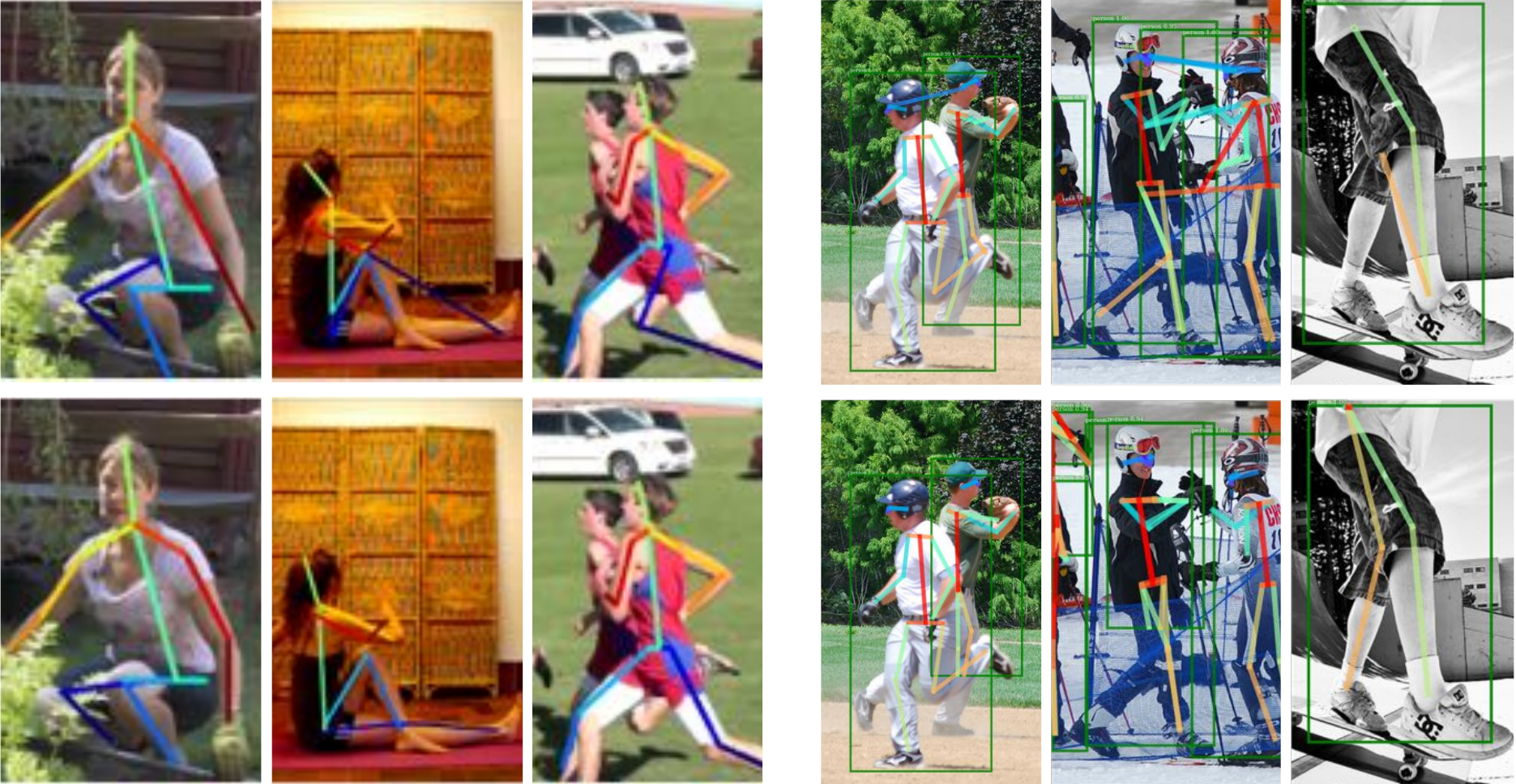}\\
\hspace*{5mm} (a)\hspace{66mm} (b)
\vspace{-.25cm}
\caption{{\small MDN-based improvements in human pose estimation through (a) within-part voting (MPII dataset, \textit{Single person} track, ResNet-152 backbone) and (b) cross-part voting (COCO dataset, Mask-RCNN backbone). Top row:  baseline performance;  bottom row: MDN-corrected pose estimates.}}
\vspace{-.5cm}
\label{fig:filters}
\end{center}
\end{figure*}

\begin{table*}[!t]
  \begin{center}
  \caption{\small Relative performance of a ResNet-152-MD network applied to cross-voting between joints.}\vspace*{3mm}
  \label{tbl:global}
    \resizebox{.58\linewidth}{!}{
    \begin{tabular}{l|c|c|cc}
      \toprule
\multirow{2}{*}{\hspace*{4mm} Model } 
&\multirow{2}{*}{\!\!\!\begin{tabular}{l} \hspace*{6pt}No\\voting \end{tabular}\!\!\!}
&\multirow{2}{*}{\!\!\!\begin{tabular}{l} Bilinear\\\hspace*{5pt}kernel \end{tabular}\!\!\!}
& \multicolumn{2}{c}{Gaussian kernel}\\ \cmidrule{4-5}
     &&& $k_f{=}5$ &
     $k_f{=}11$ \\ \midrule
Baseline, additive & \multirow{1}{*}{83.96} & 87.72 &  87.64 & 87.73 \\
MDN-additive &  & \textbf{88.05} & \textbf{88.08} & \textbf{87.83}  \\
\bottomrule
  \end{tabular}
  }
  \end{center}\vspace*{-3mm}
\end{table*}

\begin{table*}[t!] \centering
\begin{center}
\caption{\small Comparison with the state of the art frameworks on MPII Single person dataset (test set). \textit{Mean-val} denotes mean PCKh on the validation set. (*) -- models based on the Hourglass architecture.
\label{tbl:resultsmpiisp}}\vspace*{3mm}
\resizebox{.98\linewidth}{!}{
\begin{tabular}{l|cccccccc|c}
\toprule
\small \hfill Model \hfill\, & Head\!& \!\!Shoulder\!\! & \!\!Elbow\!\! & \!\!Wrist\!\! & \!\!Hip\!\! & \!\!Knee\!\! & \!\!\!Ankle\!\!\! & \!\!Mean\!\! & \!\!\textit{Mean-val}\!\!\\
\midrule
Chu et al.\cite{ChuYOMYW17}& 98.5  & 96.3  & 91.9  & 88.1  & 90.6  & 88.0 & 85.0 & 91.5$\vphantom{t}^{*}$ & 89.4$\vphantom{t}^{*}$ \\
Newell et al. \cite{NewellYD16} & 98.2  & 96.3  & 91.2  & 87.1  & 90.1  & 87.4 & 83.6 & $90.9^{*}$ & 89.4$\vphantom{t}^{*}$ \\
Bulat et al. \cite{BulatT16}& 97.9  & 95.1  & 89.9  & 85.3  & 89.4  & 85.7 & 81.7 & $89.7^{*}$ & 88.2$\vphantom{t}^{*}$\\ 
\midrule
Wei et al. \cite{WeiRKS16}& 97.8  & 95.0  & 88.7  & 84.0  & 88.4  & 82.8 & 79.4 & 88.5 & --\\
Insafutdinov et al.\cite{InsafutdinovPAA16} & 96.8  & 95.2  & 89.3  & 84.4  & 88.4  & 83.4 & 78.0 & 88.5 & --\\ 
Belagiannis et al. \cite{BelagiannisZ17}\!\!\!& 97.7  & 95.0  & 88.2  & 83.0  & 87.9  & 82.6 & 78.4 & 88.1 & 86.3\\ 
Rafi et al. \cite{RafiGL16}& 97.2  & 93.9  & 86.4  & 81.3  & 86.8  & 80.6 & 73.4 & 86.3 & --\\
Gkioxari et al.\cite{GkioxariTJ2016}& 96.2  & 93.1  & 86.7  & 82.1  & 85.2  & 81.4 & 74.1 & 86.1 & 85.3\\
Lifshitz et al. \cite{LifshitzFU16} & 97.8  & 93.3  & 85.7  & 80.4  & 85.3  & 76.6 & 70.2 & 85.0 & --\\
Pishchulin et al. \cite{deepcut}& 94.1  & 90.2  & 83.4  & 77.3  & 82.6  & 75.7 & 68.6 & 82.4 & --\\
Hu\&Ramanan \cite{HuR16}& 95.0  & 91.6  & 83.0  & 76.6  & 81.9  & 74.5 & 69.5 & 82.4 & -- \\
Carreira et al. \cite{carreiraA2016} & 95.7  & 91.7  & 81.7  & 72.4  & 82.8  & 73.2 & 66.4 & 81.3 & --\\ 
\midrule
Hourglass-8-MDN & 98.2 & 96.4 & 91.6 & 87.4 & 90.8 & 87.9 & 84.3 & 91.3$\vphantom{t}^{*}$ & 89.7$\vphantom{t}^{*}$\\ 
Resnet-152-MDN & 97.7  & 95.8  & 90.4  & 85.1  & 88.9  & 85.6 & 81.6 & 89.7 & 88.6\\
\bottomrule
\end{tabular}
}
\end{center}\vspace*{-3mm}
\end{table*}

\subsubsection{Evaluation results}



In Table \ref{tbl:local} we compare  Mass Displacement  Networks for within-part voting over a set of increasingly complicated baselines: a) a network trained with the binary cross-entropy loss with a single objective in the form of a joint heatmap, b) a network outputting the first round of posterior probabilities and displacements independently with following aggregation of corresponding votes in the form of post-processing, i.e. without end-to-end training, c) a modified Spatial Transformer network (STN) \cite{JaderbergSZK15} aiming on shrinking the produced distributions from iteration to iteration and, just as our architecture, trained end-to-end. In this case, the spatial transformation is not defined globally but instead learned in the form of a vector field describing pixel-wise linear translation.

In the top rows of Table \ref{tbl:local} we evaluate the baseline performance for different filter sizes and gauge the impact of this choice. This is easier to do since mass aggregation is done as postprocessing and does not require re-training of the model. We then train MDN models specifically with selected filter sizes.

We observe that MDNs yield  a substantial boost over the different simpler baselines, even when end-to-end training is used, as in the case of STNs. This latter aspect can be attributed to the evidence accumulation operation of MDNs, which is better suited than interpolation (STNs) for the task.\\
The support of the kernel determines the computational complexity of the MD module;  we note that by training MDNs end-to-end we achieve excellent results even with $2{\times}2$ bilinear kernels, rather than using extended Gaussian kernels. Intuitively, we train our voting network to throw more accurate shots towards the center of the geometric structures, making the use of large kernels unnecessary. 

In Table \ref{tbl:resnets} we repeat the same evaluation for different feature extractors with a varying set of network architectures -- the results indicate that there is a consistent improvement thanks to the MDN module, and that in all tasks the noisy-or and the additive voting yield virtually identical results. This confirms that we can discard the  ad-hoc choice of training the second stage with regression, and replace it with the more meaningful cross-entropy loss.  

We next evaluate  MDNs  on the task of  passing information across different joints (\textit{cross-voting}).
The corresponding results are shown in Table \ref{tbl:global}.  All models have now been trained to produce three kinds of outputs: posterior probabilities, local offsets and across-part voting offsets. This explains the drop in the baseline's performance, which was forced to  a harder multi-task learning setting (see Table \ref{tbl:local} for comparison of the single-task network performance). However, employing an MD layer in the global setting leads to substantial improvement in the localization performance. 

Comparison with the state-of-the-art methods is provided in Table~\ref{tbl:resultsmpiisp}. It shows that the MDN version of Resnet-152 outperforms all methods not based on Hourglass architecture, while Hourglass-MDN gives a 0.4 point boost over the corresponding baseline and is competitive with the most complex methods \cite{ChuYOMYW17,BulatT16}. 

Finally, our experiments have shown that stacking several mass displacement modules in different ways (within+across, across+within, as well as several modules of the same kind) does not further improve performance. This could be explained by the fact that within-part voting is included in cross-joint aggregation (each joint also votes for itself) and, at the same time, dropping across-joint connections in simple cases allows the model to focus its capacity on local aggregation more efficiently. As a result, local voting performs better in the single person setting while cross-joint scheme turned out to be most effective in the multi-person scenario.

\subsection{Multi-person pose estimation}
We have obtained similar improvements as the ones reported above also on the  challenging task of multi-person pose estimation \textit{in the wild}, which includes both object detection 
and pose estimation. We have built on the recently-introduced Mask-RCNN system of \cite{maskRCNN} which largely simplifies the task by integrating object detection and pose estimation in an end-to-end trainable architecture. This method has been shown to be only marginally inferior to two-stage architectures, like \cite{gpapan} that first detect objects, and then apply pose estimation on images cropped around the detection results. 

As in our previous experiments, we have  extended the Mask-RCNN architecture with two displacement branches ($o_x$ and $o_y$) that operate in parallel to the original classification, bounding box regression and pose estimation heads.
In the setting of \textit{cross-part} voting, we trained the whole architecture on COCO end-to-end, using identical experimental settings as those reported in \cite{maskRCNN}. As shown  in Table \ref{coco}, our MDN-based modification of Mask-RCNN yields a substantial boost in performance over the original Mask-RCNN architecture. We also obtain results that  are directly comparable to \cite{gpapan}, while employing a substantially simpler and faster architecture. 

Finally, in Appendix B we show that adding the mass displacement module with additional supervision on offsets further improves performance of detection branches (see Tables \ref{tbl:detection} and \ref{tbl:masks}).



\begin{table*}[t!]
  \begin{center}
  \caption{\small Performance of state-of-the-art pose estimation models trained exclusively on COCO data and tested on COCO \texttt{test-dev} (same as in \cite{maskRCNN}).  
\label{coco}}\vspace*{3mm}
  \resizebox{0.9999\linewidth}{!}{
    \begin{tabular}{l|ccc|cc|ccc|cc}
      \toprule
     \hfill Method \hfill\, & $\text{AP}^{\text{kp}}$\!\! & \!\!\!\!$\text{AP}^{\text{kp}}_{50}$\!\!\! & \!\!\!$\!\!\text{AP}^{\text{kp}}_{75}$\!\! & \!\!$\text{AP}^{\text{kp}}_{M}$\!\!\!\! & \!\!\!\!$\text{AP}^{\text{kp}}_{L}$\!\!  & \!\!$\text{AR}^{\text{kp}}$\!\!\! & \!\!\!$\!\!\text{AR}^{\text{kp}}_{50}$\!\! & \!\!\!\!\!\!$\text{AR}^{\text{kp}}_{75}$\!\! & \!\!$\text{AR}^{\text{kp}}_{M}$\!\!\!\! & \!\!$\!\!\text{AR}^{\text{kp}}_{L}$\!\!\\
\midrule
Mask R-CNN, keypoints \cite{maskRCNN}& 62.7 & 87.0 & 68.4 & 57.4 & 71.1 & -- & -- & -- & -- & --\\
Mask R-CNN, masks{+}keypoints \cite{maskRCNN}\!\!\! & 63.1 & 87.3 & 68.7 & 57.8 & 71.4 & -- & -- & -- & -- & -- \\
RMPE \cite{CmuPose} & 61.0 & 82.9 & 68.8 & 57.9 & 66.5 & -- & -- & -- & -- & --\\
CMU-Pose \cite{CmuPose} & 61.8 & 84.9 & 67.5 & 57.1 & 68.2 & 66.5 & 87.2 & 71.8 & 60.6 & 74.6\\
G-RMI, COCO only \cite{gpapan}& 64.9 & 85.5 & 71.3 & 62.3 & 70.0 & 69.7 & 88.7 & 75.5 & 64.4 & 77.1 \\
\midrule
Mask R-CNN-MDN, keypoints & 63.9 & 87.2 & 70.0 & 58.5 & 72.3 & 70.7 & 91.9 & 76.2 & 64.8 & 78.8\\
\bottomrule
  \end{tabular}
  }
  \end{center}
  \vspace*{-4mm}
\end{table*}


\section{Conclusion}

In this work we have introduced Mass Displacement Networks, a principled approach to integrate voting-type operations within deep architectures. MDNs provide us 
with a method to accumulate evidence from the image domain through an end-to-end learnable operation. 
We have demonstrated systematic improvements over strong baselines in human pose estimation, in both the single-person and multi-person settings. 
The geometric accumulation of evidence implemented by MDNs is generic and can  apply to  other   tasks such as surface, curve and landmark estimation in 3D volumetric data in medical imaging, or curve tracking in space and time -- we intend to explore these in the future.

\section*{Appendix A}
The voting transformation is described in \refeq{eq:voting} as follows:
\ba
m(\posout) = \sum_{\pos} K_{\sigma}(\posout  - \left[\pos + \dsp(\pos)\right]) c(\pos).
\ea
If one  interprets both 
$c(\cdot)$ and $m(\cdot)$ as fields of posterior probability values,  one has:
\ba
c(\pos)\!=\!1, ~~ \dsp(\pos)\! =\! \pos\! -\!\posout ~~ \rightarrow m(\posout)\! =\! \sum_{\pos}\! K(\pos) >1
\ea
In this case, ensuring that $m(\posout){\leq}1$
would mean  that we must use a normalized kernel, e.g.  $K(\xx){=}\frac{1}{2\pi \sigma^2} \exp(-\frac{\|\xx\|^2}{2\sigma^2})$, as used in \cite{gpapan}. 
One counter-intuitive resulting property is that the input-output mapping function defined by \refeq{eq:voting}  can result in a decrease, rather accumulation  of evidence. Consider in particular 
a perfectly-localized and perfectly-confident local evidence signal 
expressed in the form of a delta function centered at $\xx$:
\ba
c(\pos) &=& \left\{\begin{array}{cc} 1,& \pos = \xx \\ 0,& \mathrm{otherwise}\end{array}\right.
\ea
The result of voting according to \refeq{eq:voting} would then be  a blurred support map that only yields a maximal support of $\frac{1}{2\pi \sigma^2}$
to $\xx + \dsp(\xx)$:
\ba
 m(\pos) &=&  \frac{1}{2\pi \sigma^2}\exp\left(-\frac{\|\pos - (\xx + \dsp(\xx))\|^2}{2\sigma^2}\right).
\ea
For a large value of $\sigma$ this can result in an arbitrarily low value of $m(\pos)$, which is counter-intuitive, given the originally strong evidence at $\xx$. At the root of this  problem lies the operation of summing probabilities, which is a common operation when marginalizing over hidden variables, but does not  make sense as a method of accumulating evidence \cite{Williams}. 

\section*{Appendix B. }

Finally, we perform an ablation study in the multi-task setting to analyze the effect of the introduced cross-part MDN module on the performance of other brances of Mask R-CNN, namely bounding box regressor (Table \ref{tbl:detection}) and predictor of binary masks (Table \ref{tbl:masks}) for the \textit{person} class from COCO \texttt{minival}. In both cases, we observed consistent improvements in performance across the whole set of evaluation metrics.
However, in the presence of the MDN module, activating the mask branch does not further improve the quality of pose estimation as in the baseline case.

\begin{table*}[!h]
 \begin{center} 
 \caption{\small Object detection performance (bounding box AP/AR) on COCO \texttt{minival}, \textit{person} class.  
\label{tbl:detection}}\vspace*{1mm}
  \resizebox{0.99\linewidth}{!}{
    \begin{tabular}{l|ccc|cc|ccc|cc}
      \toprule
     \hfill Method \hfill\, & $\text{AP}^{\text{100}}$\!\! & \!\!$\text{AP}^{\text{100}}_{50}$\!\! & \!\!$\text{AP}^{\text{100}}_{75}$\!\! & \!\!$\text{AP}^{\text{100}}_{M}$\!\! & \!\!$\text{AP}^{\text{100}}_{L}$\!\!  & \!\!$\text{AR}^{\text{1}}$\!\! & \!\!$\text{AR}^{\text{10}}$\!\! & \!\!$\text{AR}^{\text{100}}$\!\! & \!\!$\text{AR}^{\text{100}}_{M}$\!\! & \!\!$\text{AR}^{\text{100}}_{L}$\\
\midrule
Mask R-CNN, bb& 51.5 & 82.5 & 55.0 & 59.4 & 68.5 & 18.2 & 52.2 & 59.8 & 66.9 & 76.3\\
Mask R-CNN, bb{+}mask& 52.2 & 83.1 & 55.9 & 59.8 & 69.7 & 18.4 & 52.8 & 60.4 & 66.9 & 77.2 \\
\midrule
Mask R-CNN, bb{+}keypoints		     & 51.6 & 81.4 & 55.3 & 60.1 & 69.7 & 18.3 & 52.4 & 60.0 & 67.4 & 77.0 \\
Mask R-CNN-MDN, bb{+}keypoints & 52.0 & 81.8 & 55.9 & 60.7 & 70.0 & 18.5 & 52.9 & 60.3 & 67.7 & 77.3\\
\midrule
Mask R-CNN, bb{+}mask{+}keypoints& 51.7 & 81.6 & 55.6 & 60.1 & 69.8 & 18.4 & 52.6 & 60.3 & 67.6 & 77.2 \\
Mask R-CNN-MDN, bb{+}mask{+}keypoints\! & 52.2 & 81.6 & 56.4 & 60.6 & 71.1 & 18.7 & 53.3 & 61.0 & 68.1 & 78.3\\
\bottomrule
  \end{tabular}
  }
  \end{center}
  \vspace*{-3mm}
\end{table*}

\begin{table*}[!h]
  \begin{center}
  \caption{\small Instance segmentation performance (mask AP/AR) on COCO \texttt{minival}, \textit{person} class. 
\label{masks}}\vspace*{1mm}
  \label{tbl:masks}
  \resizebox{0.97\linewidth}{!}{
    \begin{tabular}{l|ccc|cc|ccc|cc}
      \toprule
     \hfill Method \hfill\, & $\text{AP}^{\text{100}}$\!\! & \!\!$\text{AP}^{\text{100}}_{50}$\!\! & \!\!$\text{AP}^{\text{100}}_{75}$\!\! & \!\!$\text{AP}^{\text{100}}_{M}$\!\! & \!\!$\text{AP}^{\text{100}}_{L}$\!\!  & \!\!$\text{AR}^{\text{1}}$\!\! & \!\!$\text{AR}^{\text{10}}$\!\! & \!\!$\text{AR}^{\text{100}}$\!\! & \!\!$\text{AR}^{\text{100}}_{M}$\!\! & \!\!$\text{AR}^{\text{100}}_{L}$\\
\midrule
Mask R-CNN, bb{+}mask		     & 44.8 & 79.4 & 45.9 & 50.5 & 64.4 & 16.7 & 47.0 & 53.3 & 59.4 & 70.7 \\
\midrule
Mask R-CNN, bb{+}mask{+}keypoints& 45.0 & 78.5 & 47.3 & 51.4 & 65.2 & 16.8 & 47.3 & 53.8 & 60.6 & 71.4 \\
Mask R-CNN-MDN, bb{+}mask{+}keypoints & 45.6 & 78.3 & 48.1 & 52.0 & 66.1 & 17.0 & 48.1 & 54.5 & 61.3 & 72.3\\
\bottomrule
  \end{tabular}
  }
  \end{center}
  \vspace*{-3mm}
\end{table*}

\bibliographystyle{ieee}
\bibliography{egbib}

\end{document}